\pdfoutput=1
\documentclass[Journal,letterpaper]{ascelike-new}
\usepackage[utf8]{inputenc}
\usepackage[T1]{fontenc}
\usepackage{lmodern}
\usepackage{graphicx}
\usepackage[style=base,figurename=Fig.,labelfont=bf,labelsep=period]{caption}
\usepackage{subcaption}
\usepackage{amsmath}
\usepackage{csquotes}
\usepackage{newtxtext,newtxmath}
\usepackage[colorlinks=true,citecolor=red,linkcolor=black]{hyperref}
\usepackage{makecell}

\usepackage{amsthm}

\newtheorem{assumption}{Assumption}

\usepackage[linesnumbered,ruled,vlined]{algorithm2e}

\NameTag{Chen, \today}
\begin{document}
\nolinenumbers
\title{Customized Generative AI Agent for Transportation Engineering Practice: A Development and Continued Pre-training Guideline}

\author[6]{Dianwei Chen, Yuan-Zheng Lei, Zifan Zhang, Yuchen Liu, Xianfeng (Terry) Yang}




\maketitle

\begin{abstract}

Recent advancements in generative artificial intelligence (AI) and large language models (LLMs) have shown significant promise in automating complex reasoning, summarization, and question-answering tasks. However, the effectiveness of general-purpose LLMs in specialized engineering domains remains limited due to insufficient exposure to technical standards, engineering terminology, and domain-specific semantics. This study proposes a systematic approach to developing a customized generative AI agent for transportation engineering applications. A curated corpus of U.S. transportation manuals, design guidelines, and regulatory documents is used to conduct continued pretraining of six state-of-the-art LLMs through a unified low-rank adaptation (LoRA) framework. The training process is monitored to ensure convergence and model stability. Performance is evaluated using standard natural language processing metrics, including BLEU-4 and ROUGE, with Qwen2.5-7B and LLaMA-3.1-8B demonstrating the highest domain alignment and response quality. Results validate the effectiveness of LoRA-based adaptation in improving LLM performance on technical content interpretation and context-specific reasoning. This work contributes a reproducible development framework for constructing domain-specialized generative AI agents, supporting broader deployment in transportation research, design, planning, and policy analysis.

\end{abstract}
\section{Introduction}

Transportation engineering plays a vital role in enhancing the efficiency, sustainability, and livability of modern cities~\cite{vuchic2017transportation}. As urbanization accelerates, society faces growing challenges in designing transportation systems that can accommodate expanding populations while minimizing traffic congestion, reducing environmental impact, and ensuring equitable access to mobility services~\cite{pojani2015sustainable}. The transportation planning, design, and operation processes require synthesizing a wide array of information—from legal and regulatory frameworks to technical standards and engineering design guidelines—often dispersed across lengthy documents and specialized manuals~\cite{american2006planning}. For example, the Traffic Monitoring Guide issued by the U.S. Federal Highway Administration (FHWA) outlines detailed protocols for traffic data collection, sensor installation, data quality assurance, and other critical aspects of traffic monitoring~\cite{nstc2020ensuring}. While such comprehensive documents serve as valuable references~\cite{zhang2024digital}, their generality and lack of regional specificity often limit their usability in practice. Subtle but important differences in transportation policy across different areas require planners to interpret localized regulations accurately, a task that can be difficult and error-prone.


Due to the importance of the emerging technology in artificial intelligence, the main motivations of this work are listed as below three: First, the goal is to bridge the domain-knowledge gap in general-purpose LLMs by tailoring models specifically for transportation analysis. Second, this work automates the reasoning process to improve system accuracy, safety, and privacy when handling non-public documentation. Finally, a reproducible, resource-efficient fine-tuning workflow designed for the transportation research community is delivered by the proposed method.

Recent breakthroughs in natural language processing (NLP), particularly in the continued pretraining (continuing to train a pre-trained language model on additional dataset) of large language models (LLMs), have enabled transformative progress in automated text summarization, knowledge extraction, and domain-specific question-answering~\cite{zheng2023trafficsafetygpt}. Although LLMs have shown outstanding performance in legal, healthcare, and other fields~\cite{moroni2020simple,cascella2023evaluating}\, their adoption in transportation planning remains underexplored. LLMs with extensive parameterization and multi-head self-attention have demonstrated remarkable capabilities in handling complex reasoning and context-aware tasks~\cite{lavi2024fine}, suggesting that domain-specific LLMs can greatly enhance the efficiency of interpreting transportation policies, retrieving technical data, and supporting evidence-based decision-making in planning practices~\cite{cao2024large}.

Despite these advances in LLMs and other machine learning methods~\cite{chen2023using}, key challenges persist for LLM applications in transportation. Regional variation in policy documents and technical manuals can compromise the consistency of model performance. Misinterpretation of regulatory language can introduce risks in legal and compliance contexts. Additionally, growing concerns over privacy must be addressed when models handle sensitive traffic document, such as unpublicized work zone operation guide~\cite{laskar2024systematic}. 
In this study, we investigate the potential of LLMs to support the interpretation of complex transportation documents, using FHWA publications as test cases. By building an LLM-powered question-and-answer system, we demonstrate how NLP can streamline document review, improve access to policy knowledge, and assist planners in navigating the nuanced regulatory landscape across different states.


Our contributions are summarized in three folds:
\begin{itemize}
  \item We design a robust PDF-to-JSON preprocessing method that extracts and normalizes transportation policy text (titles, chapters, clauses) to form a high-quality continued-pretraining corpus.
  \item We integrate LoRA adapters into six large language models for domain-specific continued pretraining, achieving parameter efficiency by only updating low-rank matrices while preserving the original model structure.
  \item We derive a convergence guarantee for the LoRA optimization under standard L-smoothness and bounded-below assumptions, and demonstrate through experiments on the FHWA dataset that our adapted model outperforms baseline fine-tuning in both perplexity and downstream QA tasks.
\end{itemize}

The rest of the paper is organized as follows: the next section provides a comprehensive review of existing studies and discusses the current research gaps; the "Methodology" section specifies the framework that transportation engineers can follow to develop an AI agent with LLM training; the "Experiment" section conducts the model evaluation and demonstrates the effectiveness of different LLMs; and the last section summarizes the key findings.

\section{Related Works}
\subsection{Textual extraction and analysis in transportation}
Current urban planning and transportation design have been influenced by a variety of policy documents, technical manuals, and regulatory guidelines~\cite{weiner2016urban}. Traditionally, most researchers have relied on keyword extraction, searching, rule-based understanding, etc., to distill key points from a large amount of urban planning literature~\cite{puri2023commonsense}. Even though these methods can provide basic data retrieval and classification functionalities~\cite{kroeze2003differentiating,tsuji2018appling}, sometimes they struggle to deal with the detailed paragraphs and extensive domain-specific knowledge typical of transportation planning and engineering documents~\cite{mishra2017domain}.

One key limitation of traditional methods is their inability to handle the contextual complexity of transportation policies~\cite{winter2017characterizing,liu2024digital}. Many transportation regulations involve conditional clauses, interdependent rules, and calculation-needed numerical limitations that simple text-mining techniques fail to capture~\cite{massey2013automated}. For example, the standard specifics for road infrastructure may depend on factors such as traffic volume, road type, and environmental conditions. Human sense and rule-based systems often require large manual work on classification, identification, and system establishment~\cite{buhler2020analysing,chen2024deep}. This process is usually not easily adaptable when policies are updated or new guidelines are introduced.
Additionally, urban planning and technical documents~\cite{zhang2025synergizing} often include complex elements such as tables, cross-referenced annexes, and specialized legal language, which pose challenges for rule-based or simple machine learning models~\cite{moroni2020simple}. The hierarchical structure and domain-specific terminology, such as traffic light codes or transportation standards tied to transportation construction use, require a deeper contextual understanding~\cite{waltl2018rule}.

This complexity highlights the limitations of traditional methods and the need for more advanced, context-aware models. LLMs, when fine-tuned on domain-specific documentation, can better interpret complex relationships, understand minor differences in the documentation, and provide accurate summaries or policy comparisons~\cite{rasheed2024can}. 

In our work, we overcome these limitations by first converting raw PDF policy texts into a structured JSON corpus that retains hierarchical headings and conditional clauses. We then continue pretraining six base large language model with LoRA adapters on this corpus, enabling the model to internalize domain-specific rules and dependencies—tasks that keyword- or rule-based systems cannot perform without extensive manual engineering.

\subsection{Large language models in policy and technical document processing}

LLMs have demonstrated substantial potential in processing complex policy and technical documents by capturing detailed contexts and intricate dependencies that are typically challenging for traditional text-mining techniques~\cite{tang2023policygpt,karapantelakis2024using}. These models utilize deep neural network architectures trained on extensive textual datasets, enabling them to understand and generate human-like responses, summarize extensive documents, and extract critical information with high precision~\cite{chen2025insight}. When continually pretrained or fine-tuned specifically for transportation policy and technical guidelines, LLMs can significantly enhance decision-making processes and policy compliance for implementers by automating tedious understanding tasks~\cite{gunes2023multiclass}.

Recent advancements have shown that fine-tuning LLMs with domain-specific documentation, including regulatory guidelines, engineering specifications, and transportation standards, significantly improves their contextual understanding and interpretability~\cite{wandelt2024large,smetana2024highway,nikbakht2024tspec}. For example, fine-tuned LLMs can effectively handle conditional terms within policies, interpret interconnected rules across documents, and accurately extract numeric thresholds for compliance purposes. 

Implementing LLMs in transportation documentation processing offers various benefits such as reducing human error, improving efficiency, and enhancing responsiveness to policy changes~\cite{wandelt2024large}. These models also help the comparative analysis of policy variations, summarization of regulatory updates, and quick identification of discrepancies or conflicts within extensive documentation~\cite{erak2024leveraging}. Such capabilities are invaluable for urban planners and transportation engineers, empowering them to focus on strategic planning and innovation rather than routine document parsing and analysis tasks~\cite{zhai2024enhancing}.

Unlike prior approaches that fine-tune all model parameters—incurring high compute and storage costs—our framework injects lightweight LoRA adapters into baseline large language models. By freezing the dense model weights and only updating low-rank matrices, we achieve comparable or better domain performance while reducing trainable parameters by over 95\%. This makes continued pretraining feasible on larger documents and longer contexts typical of transportation policy corpora.

\subsection{Existing Practices and Challenges in Policy Document Interpretation and Q\&A Systems}

Automated interpretation of transportation policy documents using question-answering (Q\&A) systems has received increased attention in recent studies~\cite{veena2019ontology}. Several initiatives have explored using NLP techniques, particularly well-trained language models, to generate key policy requirements, regulatory standards, and implementation guidelines from transportation documents~\cite{haduong2024risks}. For instance, automated Q\&A frameworks have been successfully applied to regulatory manuals by transportation agencies to improve policy compliance checking and reduce manual review workloads~\cite{amalina2024public}. Despite these advancements, practical applications encounter significant difficulties~\cite{ramaraj2252improving,mavrogiorgos2023question}. Challenges include domain-specific policy variations leading to consistency issues~\cite{liang2024internal}, increased legal risks associated with incorrect interpretation of regulatory language, and confidentiality use requirements for transportation-related documentation before it was published, including abbreviations, annex references, and highly specialized terminology.

Our framework addresses these challenges by integrating an initial PDF cleaning pipeline that preserves and tags sensitive sections (e.g., annexes, legal clauses) before continued pretraining. The LoRA-adapted large language model then learns to disambiguate abbreviations and interpret conditional clauses without exposing raw documents, enabling secure, compliant Q\&A services with reduced risk of misinterpretation.

\section{Methodology}
\begin{figure}[htbp]
    \centering
    \includegraphics[width=1.2\linewidth]{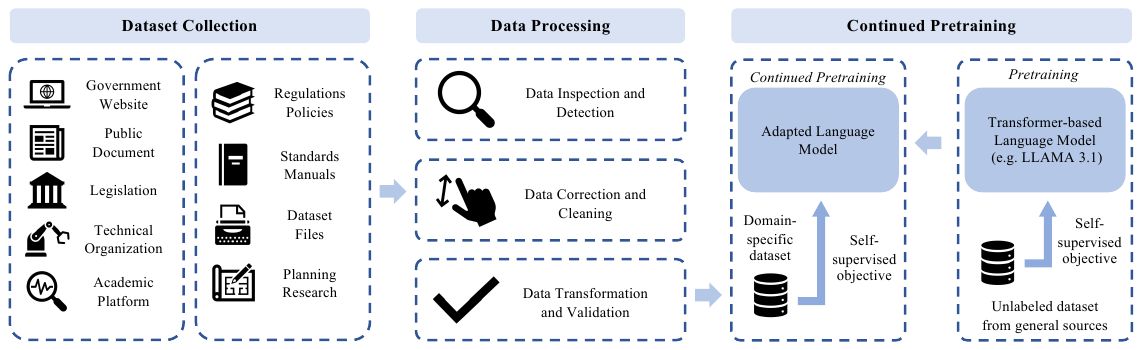}
    \caption{Transportation domain-specific continued pretraining framework}
    \label{fig:framework}
\end{figure}

As shown in Fig.~\ref{fig:framework}, our pipeline consists of three steps: (1) \emph{Document Ingestion and Cleaning}, where PDFs are parsed, denoised, and converted into a structured JSON corpus; (2) \emph{Continued Pretraining}, in which we freeze the base Llama 3.1 (and comparison models), inject LoRA adapters into each Transformer layer, and train these adapters on the cleaned corpus to capture domain semantics; and (3) \emph{Adapter Merging and Evaluation}, where the learned LoRA updates are merged back for inference and evaluated on downstream transportation QA and policy retrieval tasks.  

\subsection{Large language model selection and architecture}
Although many LLMs have emerged nowadays, we chose $Llama\,3.1$ as the baseline model and some other language models for comparison for pre-training domain-specific documents. For instance, the $Llama\,3.1$ adopted the transformer structure, which is a neural network architecture based on multi-head self-attention and position-wise feed-forward layers that processes entire sequences in parallel to capture long-range dependencies without recurrence, and specifically optimized the parameter design and training policy. Also, compared to previous LLMs, $Llama\,3.1$ will perform better in capturing long-range dependencies and contextual information, and then can understand and generate natural language more precisely.

$Llama\,3.1$ is currently one of the largest open source models, and its biggest incarnation is a dense Transformer with 405 B parameters and a context window of up to 128 K tokens. In this setting, a token is the smallest unit of text (e.g., a wordpiece or subword) that the model processes, and the context window specifies how many of these tokens can be attended to in a single pass—larger windows improve the model’s ability to handle long documents and maintain coherence over distant dependencies. Meanwhile, the total parameter count reflects the model’s capacity to learn complex patterns, but also dictates its memory footprint and compute requirements. Accordingly, we adopt the 8 B-parameter variant of Llama 3.1 as our baseline to balance domain adaptation efficacy with our available computational resources.

\subsection{Data preparation and preprocessing}
In our proposed method, the original Portable Document Format (PDF) document will be provided as first-hand documentation. As an information carrier, PDF documents will contain a large amount of unstructured or semi-structured data. 

After removing noise information such as headers, footers, page numbers, advertisements, and standardizing the format (uniform coding, punctuation, paragraphs, etc.), we are able to reduce the adverse effects of redundant information on the pre-trained model. The noise filtering rate helps to quantify the cleaning effect:
\begin{equation}
    \eta = \frac{N_{total}-N_{valid}}{N_{total}}
\end{equation}
where $N_{total}$ represents the total number of characters before text extraction and $N_{valid}$ represents the number of valid characters retained after cleaning.

The digital PDF document has preserved the structured information, such as titles, chapters, paragraphs, key clauses, etc., in the document to help subsequent text cleaning and structural processing. And also after standardizing the format of the cleaned document by using a unified encoding and removing extra spaces, line breaks, and garbled characters, the data consistency and continuity will be ensured. Finally, the PDF document will be transferred to standard JSON format, and the final pre-train dataset $P$ can be: 
\begin{equation}
    P =\left\{ \left\{''text'':N(d) \right\} | d\in D\right\}
\end{equation}
where $d$ represents the paragraphs, $N$ represents the normalization and cleaning function, and $D$ represents the PDF document.

Through the above methods, we can systematically process the original PDF data and convert it into a structured pre-trained dataset in the format of:
\begin{equation}
\begin{aligned}
     [ &\{''text'': ''document''\}, \\ & \{''text'': ''document''\},\\ & \hspace{2cm} ... \\ & \{''text'': ''document'' \}]
\end{aligned}
\end{equation}

In this research, ''Ensuring American Leadership in Automated Vehicle Technologies: Automated Vehicles 4.0''~\cite{nstc2020ensuring}, ''Preparing for the Future of Transportation: Automated Vehicles 3.0''~\cite{dot2018preparing}, and ''Manual on Uniform Traffic Control Devices for Streets and Highways $11^{th}$ Edition''~\cite{FHWA2023MUTCD11} as shown in Table~\ref{tab:dataset} has been taken as the dataset of the U.S. Department of Transportation Federal Highway Administration for the pre-training dataset. These government documents provide essential insights into U.S. AV policy, regulation, and infrastructure standards. Automated Vehicles 4.0 outlines a national strategy for AV innovation and safety. Automated Vehicles 3.0 focuses on integrating AVs into the transportation system, highlighting infrastructure and regulatory updates. The MUTCD, 11th Edition sets national standards for traffic control devices, with updates for AV compatibility. Collectively, these documents offer rich and structured textual data encompassing policy directives, engineering standards, and technical definitions. They are highly relevant for pre-training models which intended to understand the regulatory environment, infrastructure design, and operational conditions associated with automated vehicle systems in the United States.
\begin{table}[!htp]
    \centering
    \begin{tabular}{p{6cm}cp{6cm}c}
    \hline\hline
         \textbf{Documentation Name} & \textbf{Year} & \textbf{Topic} & \textbf{Page Count}\\
    \hline
         Ensuring American Leadership in Automated Vehicle Technologies: Automated Vehicles 4.0  & 2020 & U.S. government white paper that establishes guiding principles, catalogs federal agency efforts, and maps strategies to maintain American leadership in automated vehicle technologies. & 56\\
         Preparing for the Future of Transportation: Automated Vehicles 3.0   & 2018 & U.S. DOT white paper building on ADS 2.0, introducing six guiding principles for safe, technology-neutral, and consistent integration of automated vehicles across all surface transportation modes. & 80\\
         Manual on Uniform Traffic Control Devices for Streets and Highways $11^{th}$ Edition & 2023 & FHWA’s national standard detailing uniform minimum specifications and guidance for traffic signs, signals, and markings to ensure safety and efficiency on public roads. & 1161\\
    \hline
    \end{tabular}
    \caption{Training Materials for Pre-training Large Language Models}
    \label{tab:dataset}
\end{table}

\subsection{Transformer Architecture}
In this study, we take the $Llama 3.1$ language model as the baseline model for content-aware question answering~\cite{grattafiori2024llama}. The Llama model is based on the architecture of the Transformer Decoder.

The auto-regressive decoder transformer architecture is specifically made for the task generation part, such as text generation and language modeling. The feature of the auto-regressive decoder is that the next step's generated token is based on the previous one. The auto-regressive decoder is stacked with multiple decoder layers that have the same structure. The reasoning process of the auto-regressive decoder is a step-by-step, word-by-word prediction process.

Also, the multi-head attention mechanism is used in $Llama 3.1$ to allow the model to capture different levels of semantic and grammatical information and enhance its ability to model long-distance dependencies. In the autoregression decoders, multiple attention heads working together will not just increase the flexibility of the information aggregation but also make the generated text more coherent and accurate.

For example, there exists an LLM input of the inquiry from a no-knowledge-based transportation policy implementer, such as the party of the road construction area. The inquiry question '' What is meant by enhanced NHS?'' will be a series of tokens after segmentation:
\begin{equation}
   x_1:\text{"What"}  \;  x_2:\text{"is"} \; x_3: \text{"meant"} \; x_4: \text{"by"}  \;x_5: \text{"enhanced"} \; x_6: \text{ "NHS"} \; x_7:\text{"?"} 
\end{equation}
Each token corresponds to an embedding vector, forming the input matrix:
\begin{equation}
    X\in \mathbb{R}^{7\times d_{model}}
\end{equation}
where $7$ is the token number, and $d_{model}$ is the hidden dimension of the model.

The matrix $X$ will be mapped to query $Q$, key $K$, and value $V$ matrices using three linear projection matrices: 
\begin{equation}
    Q=XW_Q,K=XW_K,V=XW_V
\end{equation}
where $W_Q,W_K,W_V \in \mathbb{R^{d_{model}\times d_k}}$

For each self-attention head, the attention score of scaled dot-product attention from the transformer is a weighted sum of the values, where the weight assigned to each value is determined by the dot-product of the query with all the keys:
\begin{equation}    Attention(Q,K,V)=softmax(\frac{QK^T}{\sqrt{n}})V
\end{equation}
where the $QK^T$ represent the score $s_{ij}$ (the measure of the correlation between the $i^{th}$ token and the $j^{th}$ token in the current subspace. In this example, "What is meant by enhanced NHS?", the model will be more willing to capture the relation between the "meant" and "NHS". Also, to avoid the values of the dot-product being too large and the gradient of the softmax calculation being too small, it's necessary to divide the dot-product result by $\sqrt{d_k}$. The softmax function will be used on the scaled score matrix to get the normalized attention weight:
\begin{equation}
    a_{ij}= softmax(\hat{s}_{ij})=\frac{exp(\hat{s}_{ij})}{\sum^n_{j=1}exp(\hat{s}_{ij})}
\end{equation}
here the attention weight $a_{ij}$ represent how much does the token $i$ pay attention to the information of token $j$. The sum of each token's weight is $1$. In our proposed example, due to the feature of the auto-regression model, the token "NHS", which is compared to the token "meant", has a high score and its attention weight will be larger. The model will rely more on the information of the "NHS" when answering the question of "What is meant by enhanced NHS?"

The value matrix $V$ is the weighted sum of attention weights:
\begin{equation}
    Output_i = \sum^n_{j=1}a_{ij}v_j
\end{equation}
which represents that in a single attention head, the output of each token is obtained by weighted summing all tokens' value vector $v_j$ with the attention weight $a_{ij}$.

After the calculation of each single head and acquiring the single head attention function: 
\begin{equation}
head_i = softmax(\frac{Q_IK^T_i}{\sqrt{d_k}})V_i  
\end{equation}
by using the linear projection of each attention head: $Q_i=XW^Q_i, K_i=XW^k_i, V_i=XW^V_i$ to guarantee each head can capture different information in its subspace. Then, after concatenating the outputs of all heads through the dimension of feature to form a large vector $Concat(head_1,head_2,...,head_h) \in \mathbb{R}^{n\times(h\cdot d_k)}$ and drawing out the projection matrix $W_O \in \mathbb{R}^{(h\cdot d_k)\times d_{model} }$ to project the concatenated vector back to model's original hidden dimension $d_{model}$:
\begin{equation}
    MultiHead(Q,K,V)=Concat(head_1,head_2,...,head_h)W_O
\end{equation}

\subsection{Low-Rank Adaptation Overview}
Algorithm~\ref{alg:lora_domain} outlines a LoRA-based continued pretraining strategy for adapting a pretrained LLAMA-3.1 model to a specific domain. The process begins by freezing the original model parameters $\theta$ to preserve the foundational knowledge of the large language model. Simultaneously, low-rank adapter parameters $\Delta\theta_{\text{LoRA}}$ are initialized to enable lightweight and efficient fine-tuning. During each training epoch, the dataset $P$ is divided into mini-batches $B$, and the model iteratively processes each batch. For every input-target pair in the batch, the input is tokenized and embedded to form the input representation $X$. For each designated projection layer, LoRA is applied by injecting trainable low-rank matrices $A_i$ and $B_i$ into the corresponding weight matrix $W_i$. This modifies the attention computation, where query ($Q$), key ($K$), and value ($V$) matrices are generated from $X$, and attention outputs $Z$ are derived using scaled dot-product attention. The decoder produces a prediction $\hat{y}$ from $Z$, and the loss is calculated using the cross-entropy between the predicted and true outputs. The cumulative loss across the batch is averaged, and gradients are computed only with respect to the LoRA parameters. These parameters are updated using gradient descent with learning rate $\eta$. After training, the learned LoRA weights are merged back into the base model to form the adapted model $M'$, which is returned for domain-specific inference.

Low-Rank Adaptation (LoRA) is a parameter-efficient fine-tuning method that injects trainable low-rank matrices into existing model weights while keeping the original parameters frozen.  Given a weight matrix $W\in\mathbb{R}^{d_{\text{model}}\times d_{\text{model}}}$, LoRA introduces two smaller matrices:
\begin{equation}
A\in\mathbb{R}^{d_{\text{model}}\times r},\quad
B\in\mathbb{R}^{r\times d_{\text{model}}},
\end{equation}
with $r \ll d_{\text{model}}$, and replaces the original weight in forward computation by
\begin{equation}
W' = W + A B.
\end{equation}
During continued pretraining, only $A$ and $B$ are updated, yielding a low-rank update $\Delta W = AB$ that significantly reduces the number of trainable parameters. After training, $\Delta W$ can be merged into $W$ for efficient inference without changing the model’s original structure.

\subsection{Convergence Derivation}
We consider the optimization of the LoRA adapters $\Delta\theta=(A,B)$  by minimizing:
\begin{equation}
g(\Delta\theta) = f\bigl(\theta + A\,B\bigr), 
\end{equation}
where $\theta$ are the frozen base model parameters and $f$ is the pretraining loss.

First, the three assumptions are defined below:

\begin{assumption}[L-smoothness]
The function $g$ is differentiable, and there exists $L > 0$ such that for all $\Delta\theta, \Delta\theta'$,
\begin{equation}
    \|\nabla g(\Delta\theta) - \nabla g(\Delta\theta')\|
    \le L\,\|\Delta\theta - \Delta\theta'\|.
\end{equation}
\end{assumption}

\begin{assumption}[Lower bound]
The function $g$ is bounded below:
\begin{equation}
    g^* = \inf_{\Delta\theta} g(\Delta\theta) > -\infty.
\end{equation}
\end{assumption}

\begin{assumption}[Gradient descent]
The parameter update follows standard gradient descent with a constant step size $0 < \alpha \le 1/L$:
\begin{equation}
    \Delta\theta_{t+1}
    = \Delta\theta_t - \alpha\,\nabla g(\Delta\theta_t).
\end{equation}
\end{assumption}

These assumptions are essential for proving convergence. \textbf{L-smoothness} ensures the gradient does not change too rapidly, allowing us to bound the loss after each update. The \textbf{lower bound} guarantees that the loss cannot decrease indefinitely. The \textbf{gradient descent} rule with a proper step size ensures stable and controlled updates. 

By $L-smoothness$ and the update rule, we have
\begin{equation}
    \begin{aligned}
      g(\Delta\theta_{t+1})
      &\le
      g(\Delta\theta_t)
      + \langle \nabla g(\Delta\theta_t),\,\Delta\theta_{t+1}-\Delta\theta_t \rangle
      + \tfrac{L}{2}\|\Delta\theta_{t+1}-\Delta\theta_t\|^2 \\[4pt]
      &=
      g(\Delta\theta_t)
      - \alpha\|\nabla g(\Delta\theta_t)\|^2
      + \tfrac{L\alpha^2}{2}\|\nabla g(\Delta\theta_t)\|^2 \\[2pt]
      &\le
      g(\Delta\theta_t)
      - \tfrac{\alpha}{2}\|\nabla g(\Delta\theta_t)\|^2.
    \end{aligned}
\end{equation}

Summing from \(t=0\) to \(T-1\) gives
\begin{equation}
    g(\Delta\theta_0) - g(\Delta\theta_T)
    \ge
    \tfrac{\alpha}{2}\sum_{t=0}^{T-1}\|\nabla g(\Delta\theta_t)\|^2.
\end{equation}

Since $g(\Delta\theta_T)\ge g^*$, it's easy to obtain:
\begin{equation}
    \frac{1}{T}\sum_{t=0}^{T-1}\|\nabla g(\Delta\theta_t)\|^2
    \le
    \frac{2\bigl(g(\Delta\theta_0)-g^*\bigr)}{\alpha\,T}.
\end{equation}
The above equation prove that when the iteration time $T \rightarrow \infty$, the mean gradient norm $\frac{1}{T}\sum_{t=0}^{T-1}\|\nabla g(\Delta\theta_t)\|^2 \rightarrow 0$. Thus, the limit point of the model parameter sequence ${\Delta\theta_t}$ must be a stationary point where the gradient is zero. This theoretical result is consistent with our experimental results, as shown in Section~\ref{sec:exp}, that the model will converge as the loss function decreases to a constant value that no longer varies widely

\begin{algorithm}[htbp]
\caption{LoRA-based Continued Pretraining for Domain-specific Adaptation}
\label{alg:lora_domain}
\KwIn{Pretrained LLAMA-3.1 Model $M$ with parameters $\theta$, Dataset $P$, learning rate $\eta$, LoRA rank $r$}
\KwOut{Adapted Model $M'$ fine-tuned for domain-specific tasks}

Freeze the pretrained model parameters $\theta$\;
Initialize trainable LoRA parameters $\Delta\theta_{\text{LoRA}}$\;

\For{\textbf{each} epoch}{
    \For{\textbf{each} mini-batch $B \subset P$}{
        $\text{loss} \leftarrow 0$\;
        \For{\textbf{each} $(\text{input}, \text{target}) \in B$}{
            $X \leftarrow \text{Tokenize\_and\_Embed}(\text{input})$\;

            \For{\textbf{each} target module $W_i \in [\text{gate\_proj}, \text{down\_proj}, \text{up\_proj}, \text{q\_proj}, \text{v\_proj}, \text{k\_proj}, \text{o\_proj}]$}{
                Inject LoRA adapter matrices:\;
                \quad $W_i \leftarrow W_i + A_iB_i$, 
                where $A_i \in \mathbb{R}^{d_{\text{model}} \times r}, B_i \in \mathbb{R}^{r \times d_{\text{model}}}$\;
            }

            Compute multi-head attention:
            \[
            Q = XW_Q,\quad K = XW_K,\quad V = XW_V,\quad Z = \text{softmax}\left(\frac{QK^\top}{\sqrt{d_k}}\right)V
            \]

            Generate prediction:
            \[
            \hat{y} \leftarrow \text{Decoder}(Z)
            \]

            Compute cross-entropy loss:
            \[
            \mathcal{L} \leftarrow \text{CrossEntropy}(\hat{y}, \text{target})
            \]

            $\text{loss} \leftarrow \text{loss} + \mathcal{L}$\;
        }

        $\text{loss} \leftarrow \frac{\text{loss}}{|B|}$\;
        Backpropagate gradients w.r.t. only $\Delta\theta_{\text{LoRA}}$\;
        Update LoRA parameters:
        \[
        \Delta\theta_{\text{LoRA}} \leftarrow \Delta\theta_{\text{LoRA}} - \eta \nabla_{\Delta\theta_{\text{LoRA}}}\text{loss}
        \]
    }
}

Merge LoRA parameters into $M$ for inference:
\[
M' \leftarrow M + \Delta\theta_{\text{LoRA}}
\]

\Return{$M'$}
\end{algorithm}


\section{Experiments}
 \label{sec:exp}
In this experiment, we take the Meta Llama 3.1 collection of multilingual large language models as the baseline large language models. 
\subsection{Experiment Equipment}
The proposed experiment is conducted on a lab server running Ubuntu 20.04.6 LTS (Focal Fossa). The system is equipped with a $14^{th}$ Gen Intel Core i9-14900K processor featuring 24 cores and 32 threads. The server is equipped with an NVIDIA RTX A6000 GPU with 48 GB VRAM, utilizing driver version 535.183.01 and CUDA 12.2. This high-performance GPU is optimized for AI workloads, enabling efficient training and inference of large-scale models. The system includes 192 GB of RAM and a 4 TB NVMe SSD for primary storage, supplemented by a 14.55 TB external USB drive for dataset storage and backups.

\subsection{Hyperparameter Configuration}
Based on the experiment equipment, the Llama 3.1 (text only) 8B is chosen for continued pretraining in this experiment. The low-rank adaptation method is utilized in the continued pretraining procedure, and the parameter of LoRA is shown in Table~\ref{tab:lora_parameter}.
\begin{table}[htbp]
    \centering
    \begin{tabular}{ccc}
        \hline\hline
        \textbf{Parameter Type} & \textbf{Symbol} & \textbf{Value} \\
        \hline
        Rank & $r$ & 4 \\
        Scaling factor & $lora\_alpha$ & 16 \\
        Target module list & $target\_modules$ & Varies by model$^\dag$ \\
        Dropout rate & $lora\_dropout$ & 0.1 \\
        Bias setting & $bias$ & none \\
        \hline
        \multicolumn{3}{p{12cm}}{\footnotesize$^\dag$The specific set of target modules is defined per model (e.g., \texttt{q\_proj}, \texttt{o\_proj}, \texttt{gate\_proj}, etc.) as shown in the training script.} \\
        \hline\hline
    \end{tabular}
    \caption{LoRA training configuration used in continued pretraining}
    \label{tab:lora_parameter}
\end{table}

To pre-process the dataset parameters during training, we set the maximum text sequence length to 2048 tokens as a trade-off between reducing memory usage and accelerating training speed, versus improving the model's capacity to capture long-range contextual dependencies. Also, the max sample size is set to 1000, which represents that the training dataset can choose no more than 1000 samples to ensure that the batch size does not exceed expectations due to the sample size being too large. The number of process workers for data preprocessing (such as word segmentation, truncation, etc.) is 18.

In this study, we configure the training process using Hugging Face's Training Arguments as summarized in Table~\ref{tab:train_hyperparams}. To accommodate long-sequence inputs and large model sizes, we set the per-device training batch size to 1 and apply gradient accumulation with a step size of 8 to simulate a larger effective batch size. The learning rate is initialized at $1\times10^{-4}$, a conservative value often used for fine-tuning large-scale language models. A cosine learning rate scheduler is adopted with a warm-up ratio of 0.1, allowing the learning rate to increase gradually during the initial phase of training. The model is trained for 4 epochs in total. We enable mixed-precision training (fp16) to reduce memory usage and improve computation efficiency. Logging, evaluation, and checkpoint saving are all performed every 500 steps for consistent monitoring and reproducibility. Additionally, we set a high DDP timeout value of $1.8\times10^{8}$ to ensure robust distributed training in long-running jobs.

\begin{table}[h]
\centering
\begin{tabular}{ccp{9cm}}
\hline\hline
\textbf{Parameter} & \textbf{Value} & \textbf{Description} \\
\hline
\texttt{per\_device\_train\_batch\_size} & 1 & Batch size per GPU. A small batch size helps reduce memory usage for large models or long input sequences. \\

\texttt{gradient\_accumulation\_steps} & 16 & Accumulate gradients over 16 steps before performing a parameter update, effectively simulating a larger batch size. \\

\texttt{learning\_rate} & $1\times10^{-4}$ & Initial learning rate for the optimizer; a conservative value suitable for fine-tuning large models. \\

\texttt{num\_train\_epochs} & 8 & Total number of epochs for training (i.e., full passes through the dataset). \\

\texttt{lr\_scheduler\_type} & cosine & Uses a cosine decay learning rate scheduler to gradually reduce the learning rate during training. \\

\texttt{warmup\_ratio} & 0.1 & Fraction of steps used for learning rate warm-up at the start of training. \\

\texttt{torch\_dtype} & bfloat16 & Mixed-precision training using Brain Float 16 (bfloat16) to reduce memory usage and improve computational efficiency. \\

\texttt{logging\_steps} & 1 & Logs training metrics (e.g., loss, learning rate) after every single training step. \\

\texttt{save\_steps} & 500 & Saves a model checkpoint every 500 steps for recovery and versioning. \\

\texttt{eval\_steps} & 500 & Evaluates the model on the validation set every 500 training steps. \\

\texttt{ddp\_timeout} & 180000000 & Timeout (in seconds) for distributed data parallel initialization and communication; set high to support long-running jobs. \\
\hline
\end{tabular}
\caption{Updated training hyperparameters for continued pretraining}
\label{tab:train_hyperparams}
\end{table}

\subsection{Training Result}
\subsubsection{Training Process}
As a comparison of the proposed continued pretraining method, 6 models (Baichuan-7B~\cite{hendrycks2021measuringmassivemultitasklanguage}, BLOOMZ-7B1~\cite{muennighoff2022crosslingual}, Qwen2.5-7B~\cite{yang2024qwen2}, Phi-3.5 Mini Instruct~\cite{abdin2024phi}, LLAMA-3.1-8B, Mistral-7B v0.1~\cite{jiang2023mistral7b}) are taken as the example for the experiment, as shown in Table~\ref{tab:models}. 
\begin{table}[htbp]
\centering

\begin{tabular}{>{\centering\arraybackslash}p{2cm}
                >{\centering\arraybackslash}p{2cm}
                >{\centering\arraybackslash}p{2cm}
                >{\centering\arraybackslash}p{2.5cm}
                >{\centering\arraybackslash}p{5cm}}
\hline\hline
\textbf{Model Name} & \textbf{Parameters} & \textbf{Context Length} & \textbf{Architecture} & \textbf{LoRA Target Modules} \\
\hline
Baichuan-7B & 7B & 4096 & Transformer & \texttt{W\_pack, o\_proj} \\

BLOOMZ-7B1 & 7.1B & 2048 & BLOOM & \texttt{query\_key\_value, dense} \\

Qwen2.5-7B & 7B & 32K & Decoder-only Transformer & \texttt{q\_proj, k\_proj, v\_proj, o\_proj, gate\_proj, up\_proj, down\_proj} \\

Phi-3.5 Mini Instruct & 3.8B & 128K & Decoder-only Transformer & \texttt{q\_proj, k\_proj, v\_proj, o\_proj} \\

LLAMA-3.1-8B & 8B & 8192	& Decoder-only Transformer	& \texttt{gate\_proj, down\_proj, up\_proj, q\_proj, v\_proj, k\_proj, o\_proj}\\

Mistral-7B v0.1 & 7.3B & 4096 & Transformer with GQA \& SWA & \texttt{q\_proj, k\_proj, v\_proj, o\_proj, gate\_proj, up\_proj, down\_proj} \\
\hline
\end{tabular}
\caption{Models for the experiment training}
\label{tab:models}
\end{table}

\begin{figure}[htbp]
    \centering
    \includegraphics[width=0.5\linewidth]{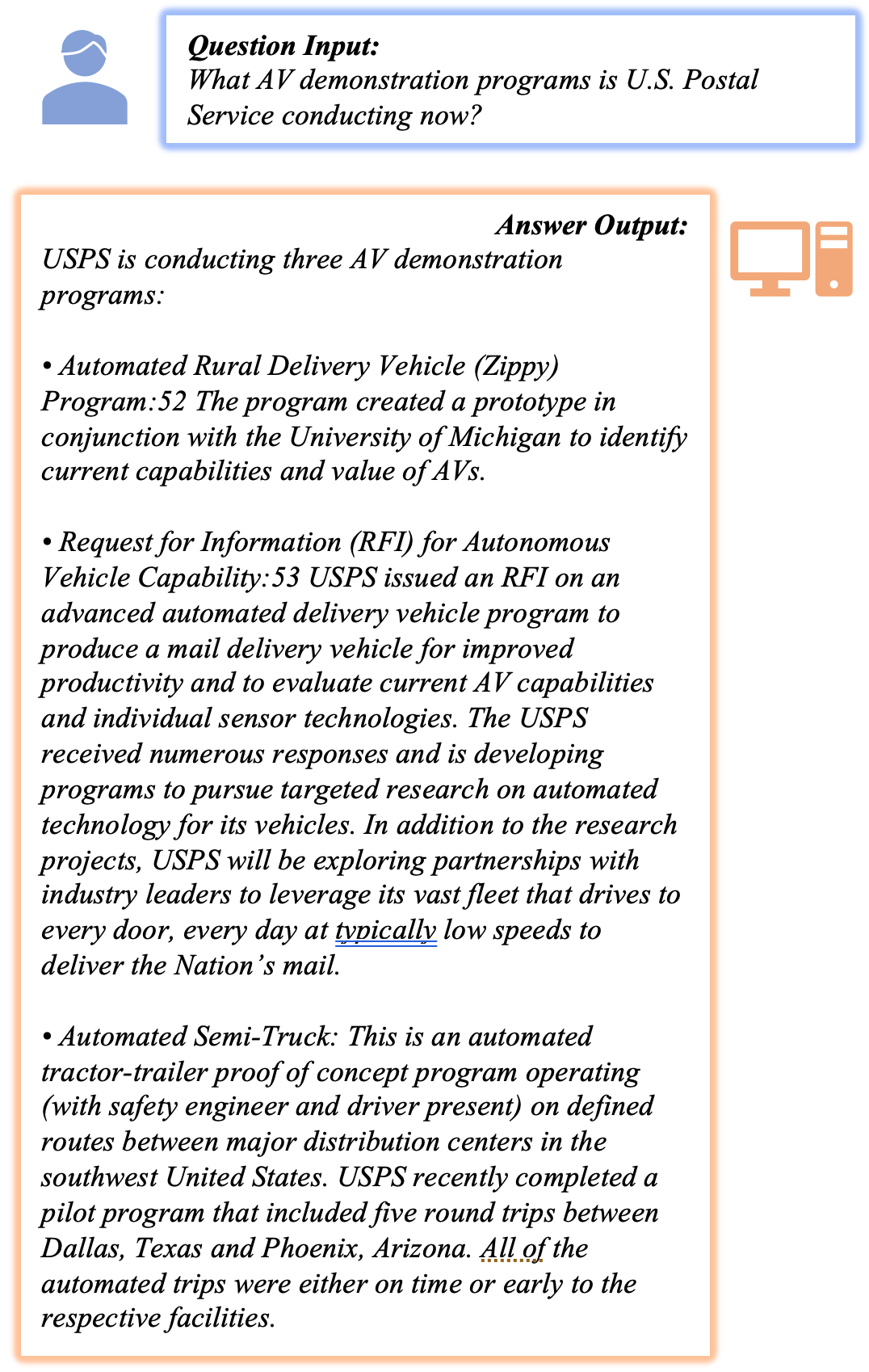}
    \caption{Demonstration of interaction with the well-trained LLMs}
    \label{fig:inter}
\end{figure}

\begin{figure}[htbp]
    \centering
    \begin{subfigure}[b]{0.8\linewidth}
        \centering
        \includegraphics[width=\linewidth]{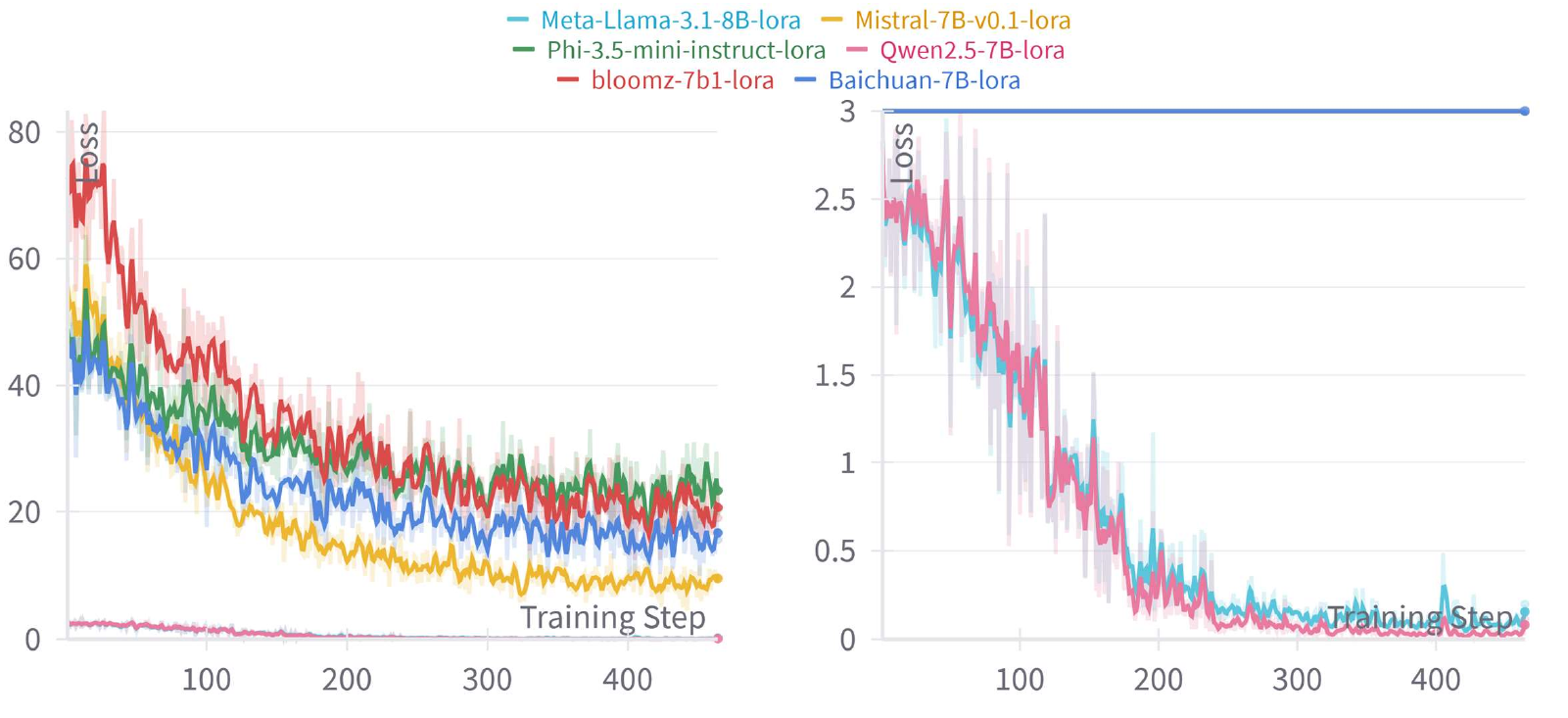}
        \caption{Training loss over epochs.}
        \label{fig:loss}
    \end{subfigure}
    \hfill


    \begin{subfigure}[b]{0.8\linewidth}
        \centering
        \includegraphics[width=\linewidth]{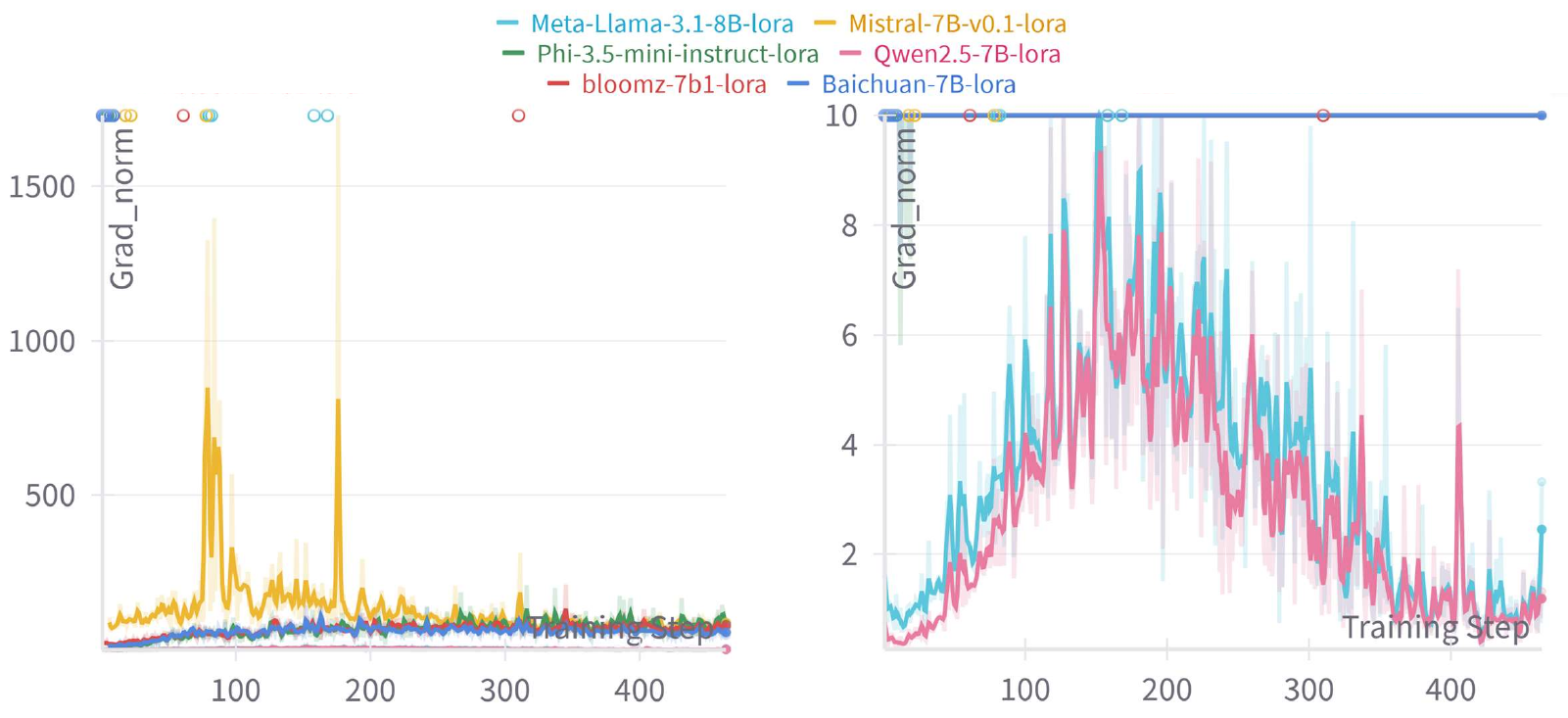}
        \caption{Gradient norm throughout training.}
        \label{fig:grad-norm}
    \end{subfigure}
    \hfill
    \begin{subfigure}[b]{0.4\linewidth}
        \centering
        \includegraphics[width=\linewidth]{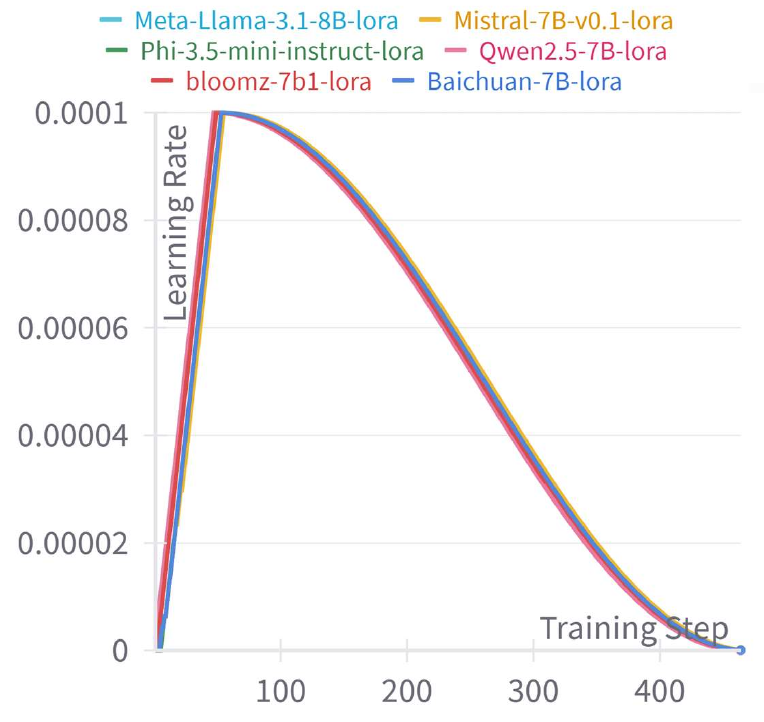}
        \caption{Learning rate schedule.}
        \label{fig:lr}
    \end{subfigure}

    \caption{Training process visualization: (a) loss curve, (b) gradient norm, and (c) learning rate schedule.}
    \label{fig:lora_training_process}
\end{figure}

The pretraining process is monitored by critical metrics such as training loss, learning rate, and gradient norm to ensure effective convergence over the domain-specific documentation. Fig.~\ref{fig:lora_training_process} illustrates the key training curves: the training loss shows a continuous decline, the learning rate follows a cosine decay schedule, and the gradient norm remains controlled, reflecting stable adaptation of the model parameters. And the final well-trained model's interaction example is shown in Fig.~\ref{fig:inter}.

During pretraining, the loss $\mathcal{L}$ of \textbf{BLOOMZ-7B1}, \textbf{Mistral-7B v0.1}, \textbf{Phi-3.5 Mini Instruct}, and \textbf{Baichuan-7B} rapidly decrease from an initial value of approximately $60$  to around $30$ within the first few hundred steps. Also, the loss $\mathcal{L}$ of \textbf{LLAMA-3.1-8B} and \textbf{Qwen2.5-7B} rapidly decrease from an initial value of approximately $2.5$  to around $0.1$ within the first few hundred steps. This sharp descent indicates that the LoRA adaptation successfully extracts the domain-specific patterns from the provided domain-specific documents. The learning rate $\eta$, initialized at $1.0 \times 10^{-4}$, decays gradually following a cosine annealing schedule. This scheduling strategy fosters smooth optimization and prevents abrupt weight updates, contributing to the overall robustness of the training. Additionally, though there exist explosion of the gradient norm from \textbf{Mistral-7B v0.1}, the gradient norm $\|g\|$ of \textbf{BLOOMZ-7B1}, \textbf{Mistral-7B v0.1}, \textbf{Phi-3.5 Mini Instruct}, and \textbf{Baichuan-7B} shows a trend of controlled reduction around $50$ and the gradient norm $\|g\|$ of\textbf{LLAMA-3.1-8B} and \textbf{Qwen2.5-7B} shows a trend of controlled reduction around $5$, suggesting that the gradient clipping and accumulation strategies are effectively preserving training stability. 

\subsubsection{Core Performance Indicators}
The continued pretrained models' performance is evaluated using several core indicators, including BLEU-4, ROUGE-L, ROUGE-1, and ROUGE-2. The BLEU-4 measures the degree of overlap between generated text and the reference text
The model's adaptation performance is further evaluated using several key indicators. ROUGE-1 captures unigram-level overlap, reflecting how well the model includes key individual words from the reference text, which is essential for evaluating basic content coverage in summarization tasks. ROUGE-2 extends this evaluation to bigrams, offering a more nuanced measure of the model’s ability to preserve local context and phrase-level coherence. ROUGE-L measures the longest common subsequence between the generated and reference texts, emphasizing the model’s ability to retain the overall structure and sequence of critical information. Thus, all four above terms provide a robust indicator of training results as shown in Table~\ref{tab:coremetric}.
\begin{table}[htbp]
\centering
\begin{tabular}{p{4.2cm}cccc}
\hline\hline
\textbf{Model Variant} & \textbf{BLEU-4 (\%)} & \textbf{ROUGE-1 (\%)} & \textbf{ROUGE-2 (\%)} & \textbf{ROUGE-L (\%)} \\
\hline
\textbf{Baichuan-7B} \\
\quad Original              & 7.3902  & 22.939 & 4.5436 & 13.939 \\
\quad Continued Pretrained  & 9.3580  & 28.792 & 9.1356 & 18.813 \\
\textbf{BLOOMZ-7B1} \\
\quad Original              & 4.3368  & 21.237 & 4.1376 & 13.409 \\
\quad Continued Pretrained  & 8.5295  & 27.882 & 8.3493 & 18.597 \\
\textbf{Qwen2.5-7B} \\
\quad Original              & 6.9313  & 23.694 & 4.5537 & 14.291 \\
\quad Continued Pretrained  & 58.083  & 68.941 & 64.725 & 65.221 \\
\textbf{Phi-3.5 Mini Instruct} \\
\quad Original              & 7.1376  & 22.946 & 4.0437 & 13.743 \\
\quad Continued Pretrained  & 7.9544  & 24.861 & 5.1860 & 15.071 \\
\textbf{LLAMA-3.1-8B} \\
\quad Original              & 7.9307  & 23.326 & 4.9744 & 14.396 \\
\quad Continued Pretrained  & 57.944  & 68.920 & 64.971 & 65.348 \\
\textbf{Mistral-7B v0.1} \\
\quad Original              & 8.2576  & 23.992 & 4.9330 & 14.593 \\
\quad Continued Pretrained  & 46.375  & 66.862 & 60.143 & 61.975 \\
\hline\hline
\end{tabular}
\caption{Comparison of core performance metrics (BLEU-4 and ROUGE) between original and continued-pretrained models on the domain-specific dataset.}
\label{tab:coremetric}
\end{table}

Table~\ref{tab:coremetric} presents the performance of six pretrained models across these four metrics. Notably, Qwen2.5-7B and LLAMA-3.1-8B significantly outperform other models in all metrics. Qwen2.5-7B achieves the highest BLEU-4 score of 58.083\%, and a ROUGE-L score of 65.221\%, indicating strong syntactic alignment and sequence-level fidelity in generated outputs. Similarly, LLAMA-3.1-8B attains competitive performance with 57.944\% BLEU-4 and the highest ROUGE-L score at 65.348\%.

In contrast, models such as Baichuan-7B, BLOOMZ-7B1, and Phi-3.5 Mini Instruct show notably lower scores. For instance, Baichuan-7B only scores 9.358\% on BLEU-4 and 18.813\% on ROUGE-L, suggesting limited capability in reproducing accurate and contextually aligned text after continued pretraining on the domain-specific dataset. Phi-3.5 Mini Instruct also stayed behind with the lowest ROUGE-2 (5.186\%) and ROUGE-L (15.071\%) scores among all models.%

Interestingly, Mistral-7B v0.1 presents a strong middle-ground performance, with BLEU-4 at 46.375\% and ROUGE-L at 61.975\%, indicating a well-balanced ability to generate text that aligns both locally and globally with the reference content.

In summary, the combination of BLEU-4, ROUGE-1, ROUGE-2, and ROUGE-L offers a comprehensive evaluation of both precision and similarity in generated outputs. As shown in Table~\ref{tab:coremetric}, models like Qwen2.5-7B and LLAMA-3.1-8B demonstrate superior adaptation through continued pretraining, while others reveal areas for further improvement.

\section{Conclusions}

This study presents a systematic investigation into the application of continued pretraining on large language models (LLMs) for domain-specific tasks in transportation policy interpretation. Using Meta's LLAMA 3.1 as the baseline for comparison, six different LLMs—including Qwen2.5-7B, LLAMA-3.1-8B, Mistral-7B v0.1, Baichuan-7B, BLOOMZ-7B1, and Phi-3.5 Mini Instruct are pretrained under a unified low-rank adaptation (LoRA) framework on a series of documents from the U.S. transportation regulatory. The training process was rigorously monitored via loss curves, learning rate schedules, and gradient norms to ensure training stability and convergence.

The models using key NLP metrics, which include BLEU-4, ROUGE-1, ROUGE-2, and ROUGE-L are evaluated. Together, these metrics give a thorough view of how closely the generated outputs match the reference texts in terms of word choice, local context, and overall structure. Among all models, Qwen2.5-7B and LLAMA-3.1-8B exhibited superior adaptation performance, achieving BLEU-4 scores above 57\% and ROUGE-L scores exceeding 65\%, clearly demonstrating their effectiveness in capturing domain-specific linguistic patterns. In contrast, models such as Phi-3.5 Mini Instruct and Baichuan-7B exhibited limited improvements, underscoring the importance of both model architecture and pretraining quality in downstream domain adaptation.

These results affirm the viability of using continued pretraining with LoRA as an efficient and scalable method to adapt general-purpose LLMs to specialized domains such as transportation planning. Moreover, the successful extraction and interpretation of complex regulatory content help develop the intelligent question-answering systems that support evidence-based decision-making in urban infrastructure design and policy compliance. Future work will further explore the integration of these domain-adapted LLMs into interactive planning tools and address challenges related to model explainability, regional bias, and data privacy.

\bibliography{ascexmpl-new}

\end{document}